\documentclass[lettersize,journal]{IEEEtran}
\usepackage{amsmath,amsfonts}
\usepackage{algorithmic}
\usepackage{algorithm}
\usepackage{array}
\usepackage[caption=false,font=normalsize,labelfont=sf,textfont=sf]{subfig}
\usepackage{textcomp}
\usepackage{stfloats}
\usepackage{url}
\usepackage{verbatim}
\usepackage{graphicx}
\usepackage{booktabs}

\usepackage{hyperref}
\usepackage[capitalize]{cleveref}
\crefname{section}{Sec.}{Secs.}
\Crefname{section}{Section}{Sections}
\Crefname{table}{Table}{Tables}
\crefname{table}{Tab.}{Tabs.}

\usepackage{multirow}
\usepackage{tabularx}
\newcolumntype{L}[1]{>{\raggedright\arraybackslash}p{#1}}
\newcolumntype{C}[1]{>{\centering\arraybackslash}p{#1}}
\newcolumntype{R}[1]{>{\raggedleft\arraybackslash}p{#1}}

\def\BibTeX{{\rm B\kern-.05em{\sc i\kern-.025em b}\kern-.08em
    T\kern-.1667em\lower.7ex\hbox{E}\kern-.125emX}}
\usepackage{balance}

\begin{document}
\title{Hierarchical Semantic Tree Concept Whitening for Interpretable Image Classification}
\author{
Haixing Dai*, Lu Zhang*, Lin Zhao, Zihao Wu, Zhengliang Liu, David Liu, \\ Xiaowei Yu, Yanjun Lyu, Changying Li, Ninghao Liu, Tianming Liu, Dajiang Zhu.
\thanks{* Co-first authors.\\
Haixing Dai, Lin Zhao, Zihao Wu, Zhengliang Liu, Ninghao Liu, Tianming Liu and Changying Li are with the
Department of Computer Science, University of Georgia, Athens, GA, USA.
(e-mail: {hd54134, lin.zhao, zw63397,zl18864, ninghao.liu, tliu}@uga.edu, changyingl@gmail.com).\\
Lu Zhang, Xiaowei Yu, Yanjun Lyu and Dajiang Zhu are with the Department of Computer Science
and Engineering, The University of Texas at Arlington, Arlington, TX, USA.
(e-mail: lu.zhang2, xxy1302, yxl9168@mavs.uta.edu, dajiang.zhu@uta.edu)\\
David Weizhong Liu is with Athens Academy, Athens, GA, USA.(e-mail:
david.weizhong.liu@gmail.com)}}

\maketitle

\begin{abstract}
With the popularity of deep neural networks (DNNs), model interpretability is becoming a critical concern. Many approaches have been developed to tackle the problem through post-hoc analysis, such as explaining how predictions are made or understanding the meaning of neurons in middle layers. Nevertheless, these methods can only discover the patterns or rules that naturally exist in models. In this work, rather than relying on post-hoc schemes, we proactively instill knowledge to alter the representation of human-understandable concepts in hidden layers. Specifically, we use a hierarchical tree of semantic concepts to store the knowledge, which is leveraged to regularize the representations of image data instances while training deep models. The axes of the latent space are aligned with the semantic concepts, where the hierarchical relations between concepts are also preserved. Experiments on real-world image datasets show that our method improves model interpretability, showing better disentanglement of semantic concepts, without negatively affecting model classification performance.
\end{abstract}


\section{Introduction}
\IEEEPARstart{M}{achine} learning interpretability has recently received considerable attention in various domains~\cite{du2019techniques, 2de2018agriculture, Murdoch-etal19defMethApp,koh2020concept}. An important challenge that arises with deep neural networks (DNNs) is the opacity of semantic meanings of data representations in hidden layers. Several types of methods have been proposed to tackle the problem. First, recent works have shown that some neurons could be aligned with certain high-level semantic patterns in data~\cite{olah2017feature,zhou2018interpreting}. Second, it is possible to extract concept vectors~\cite{Kim-etal18concepts} or clusters~\cite{Ghorbani-etal19automaticConcept} to identify semantic meanings from latent representations. However, these methods are built upon the assumption that semantic patterns are already learned by DNNs, and the models would admit the post-hoc method of a specific form. There is no guarantee that the assumption holds true for any model, especially when meaningful patterns or rules may not be manifested in the model, thus leading to over-interpretation~\cite{1rudin2019stop, Murdoch-etal19defMethApp}. Meanwhile, although many post-hoc explanation methods are proposed with the expectation of improving or debugging models, it is challenging to achieve this goal in practice. Although we could collect human annotations to guide prediction explanations and improve model credibility~\cite{Wang-etal18credibleModel, chang2021towards}, manually labeling or checking semantic concepts is rather difficult. Unlike explaining individual predictions, which is a local and instance-level task, extracting concepts provides a global understanding of models, where manual inspection of such interpretation is time-consuming and much harder, if not impossible.

Instead of relying on post-hoc approaches, we aim to instill interpretability as a constraint into model establishment. For example, explanation regularization is proposed in~\cite{ross2018improving}, but it constrains gradient magnitude instead of focusing on semantic concepts. Meanwhile, $\beta$-VAE and its variants~\cite{Higgins-etal17betaVAE, chen2019isolating} add independence constraints to learn disentangled factors in latent representations, but it is difficult to explicitly specify and align latent dimensions with semantic meanings. Ideally, we want to construct DNNs whose latent space could tell us how it is encoding concepts. The recent decorrelated batch normalization (DBN) method~\cite{12huang2018decorrelated} normalizes representations, providing an end-to-end technique for manipulating representations, but it is not directly related to interpretability.

\begin{figure}[t]
\vskip 0.2in
\begin{center}
\centerline{\includegraphics[width=\columnwidth]{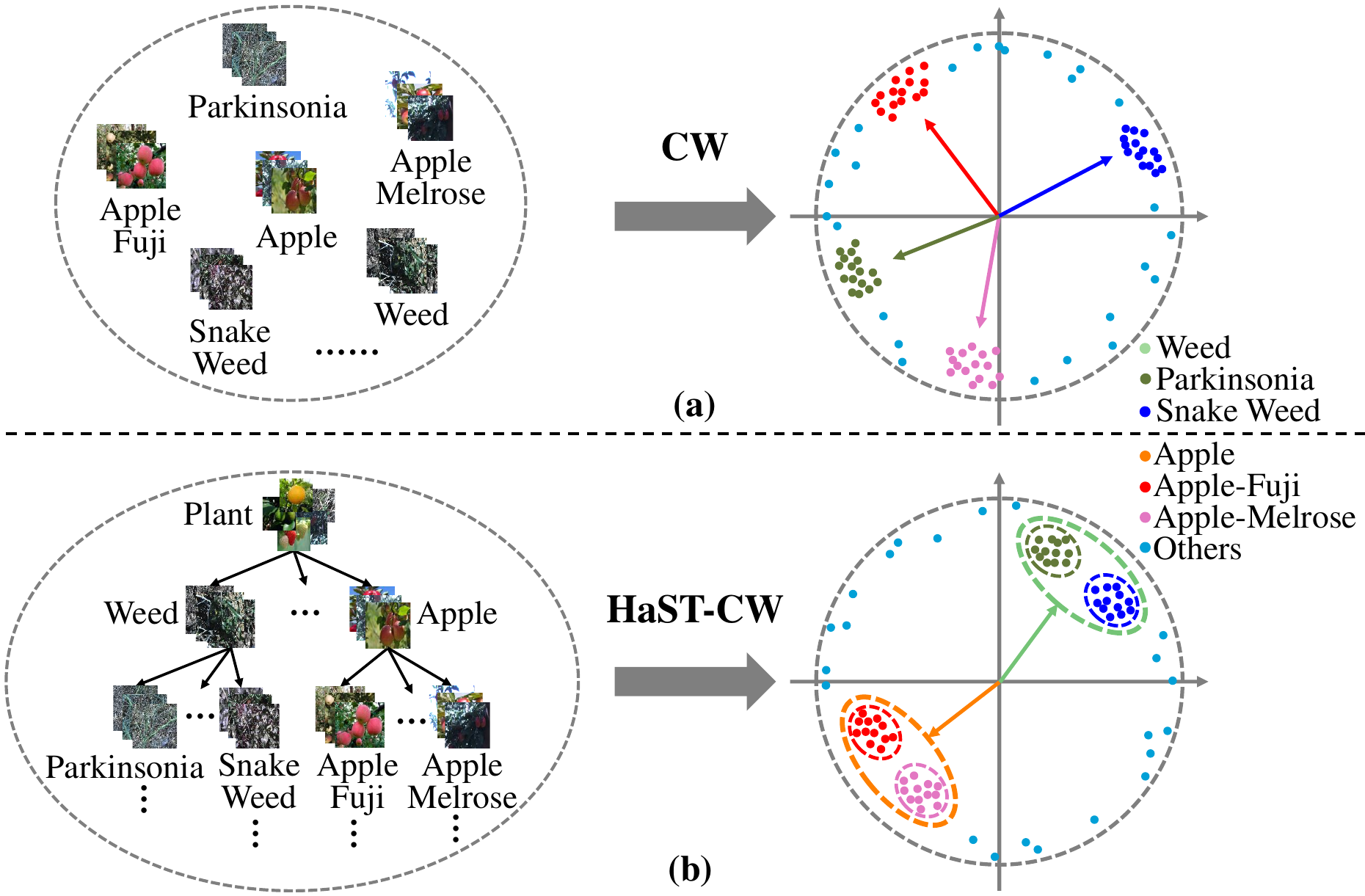}}
\caption{The intuition behind HaST-CW. (a) Distribution of discrete concepts in the latent space after applying concept whitening. (b) Distribution of hierarchical concepts after applying HaST-CW.
}
\label{fig:motivation}
\end{center}
\vskip -0.2in
\vspace{-5pt}
\end{figure}

In this work, we propose a novel Hierarchical Semantic Tree Concept Whitening (HaST-CW) model to decorrelate the latent representations in image classification for disentangling concepts with hierarchical relations. The idea of our work is illustrated in Fig.~\ref{fig:motivation}. Specifically, we define each concept as one class of objects, where the concepts are of different granularities and form a hierarchical tree structure. We decorrelate the activations of neural network layers, so that each concept is aligned with one or several latent dimensions. Unlike the traditional DBN method (Fig.~\ref{fig:motivation}a), which treats different concepts as independent, our method is able to leverage the hierarchically related organization of label concepts inherent in domain knowledge (Fig. \ref{fig:motivation}b). 
The consideration of relations between different concepts is crucial in many real-world applications~\cite{rezayi2022clinicalradiobert,rezayi2022agribert}. For example, in the healthcare domain, the relationship of different disease stages (concepts) may reflect the progression of the disease, which is significant for reversing pathology~\cite{14zhang2021representing,15zhang2020jointly,16wang2020learning,dai2023ad}. Also, in the precision agriculture domain~\cite{2de2018agriculture,rezayi2023exploring}, real-time monitoring of interactions of multiple agricultural objects (concepts) with each other and with the environment are crucial in maintaining agro-ecological balance~\cite{2de2018agriculture}. 
In our model, a novel semantic constraint (SC) loss function is designed to regularize representations. As a result, the data representations of two concepts with higher semantic similarity will be closer with each other in the latent space. Moreover, a new hierarchical concept whitening (HCW) method is proposed to decorrelate different label concepts hierarchically. We evaluated the proposed HaST-CW model using a novel agriculture image dataset called Agri-ImageNet. The results suggest that our model could preserve the semantic relationship between the label concepts, and provide a clear understanding of how the network gradually learns the concept in different layers, without hurting classification performance.

\section{Related Work}
\noindent \textbf{Post-Hoc Interpretation.} Post-Hoc interpretation can be divided into approaches that explain predictions or models~\cite{du2019techniques, Murdoch-etal19defMethApp}. Prediction-oriented interpretation aims to develop faithful and robust measures to quantify feature importance towards individual predictions for identifying those features (e.g., pixels, super-pixels, words) that made most contributions~\cite{Ribe-etal16whyshould, Smilkov-etal18smoothgrad, Selvaraju-etal16gradCAM, Bach-etal15onPixelwise, Lundberg-etal17SHAP, Ghorbani-etal19fragile}. Model-oriented interpretation analyzes behaviors of neural networks either by characterizing the function of model components~\cite{Simonyan-etal13deepInsideCNNsaliency,olah2017feature, 3zeiler2014visualizing} or analyzing semantic concepts from latent representations~\cite{Kim-etal18concepts, Ghorbani-etal19automaticConcept, bau2018identifying, mu2020compositional}. The proposed method also targets concept-level interpretation in deep neural networks. Different from post-hoc techniques that focus on discovering existing patterns in models, the newly proposed HaST-CW proactively injects concept-related knowledge into training and disentangles different concepts to promote model interpretability.

\noindent \textbf{Inherently Interpretable Models.} Another school of thought favors building inherently explainable machine learning models~\cite{rudin2019stop, chen2020concept}. Some approaches design models that highlight prototypical features of samples as interpretation. For example, Chen et al.~\cite{chen2018looks} classifies images by dissecting images into parts and comparing these components to similar prototypes towards prediction. Li et al.~\cite{li2018deep} designs an encoder-decoder framework to allow comparisons between inputs and the learned prototypes in latent space. Some other works such as $\beta$-VAE and its variants~\cite{Higgins-etal17betaVAE, chen2019isolating} regularize representation learning for autoencoders to produce disentangled factors in representation dimensions, but the semantic meaning of each dimension remains unknown without further manual inspection. In contrast, our method attempts to explicitly align latent dimensions with specific semantic concepts contained in external knowledge. A recent technique called Concept Whitening (CW)~\cite{chen2020concept} constrains the latent space, after revising Batch Whitening ~\cite{huang2018decorrelated, siarohin2018whitening}, such that it aligns with predefined classes. Our method attempts to infuse more complex knowledge of concept relations into representation learning.

\noindent \textbf{Applying Whitening to Computer Vision.}  Whitening is a standard image preprocessing technique, which refers to transforming the covariance matrix of input vectors into the identity matrix. In fact, the well-known Batch Normalization~\cite{13ioffe2015batch} can be regarded as a variant of whitening where only the normalization process is retained. There are many works in deep learning that describe the effectiveness of whitening~\cite{cogswell2015reducing, pal2016preprocessing, luo2017learning} and the process of finding the whitening matrix~\cite{desjardins2015natural}. Our work further takes semantics into consideration during the whitening process towards more interpretable representation learning.

\section{Methodology}
\subsection{Overview}
The proposed HaST-CW model aims to preserve the underlying hierarchical relationship of label concepts, as well as to disentangle these concepts by decorrelating their latent representations. To achieve this goal, we leverage the hierarchical tree structure of the label concepts extracted from specific domain knowledge (\cref{sec:Hierarchical Tree Structure of Concepts}). Then, the obtained structure of label concepts is used as prior knowledge to be instilled into the model for guiding the representation learning process. There are two key components in the knowledge instillation process -- the hierarchical concept whitening (HCW) module and the semantic constraint (SC) loss, which will be elaborated in \cref{sec:Hierarchical Concept Whitening} and \cref{sec:Semantic Constraint Loss}, respectively.

\begin{figure}[ht]
\vskip 0.2in
\begin{center}
\centerline{\includegraphics[width=0.9\columnwidth]{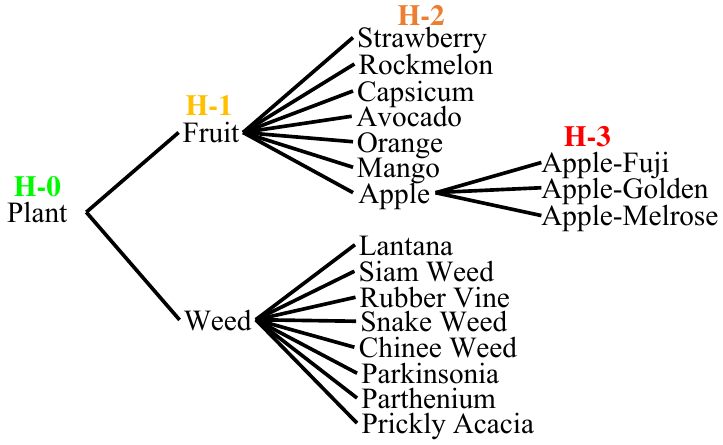}}
\caption{Hierarchical Tree Structure of Concepts.}
\label{fig:cw_tree}
\end{center}
\vskip -0.2in
\end{figure}

\subsection{The Hierarchical Semantic Tree of Concepts}
\label{sec:Hierarchical Tree Structure of Concepts}
\begin{figure}[t]
\vskip 0.2in
\begin{center}
\centerline{\includegraphics[width=\columnwidth]{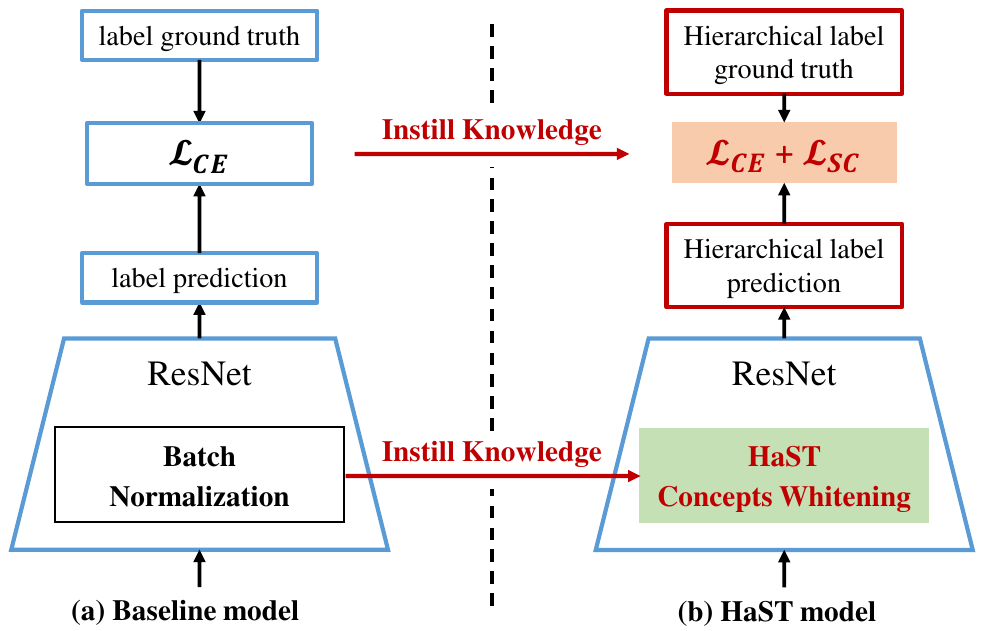}}
\caption{The architecture of HaST-CW model.}
\label{fig:model_loss}
\end{center}
\vskip -0.2in
\end{figure}

In this work, we used a newly collected and curated Agri-ImageNet dataset to develop and evaluate the HaST-CW model. There are 9173 high quality images in Agri-ImageNet, covering 21 different types of agricultural objects. Taking each type of agricultural object as one class, we have 21 label concepts in total. Some pairs of agriculture objects have the supertype-subtype relationship between them, so we obtain the parent-child relationship between the corresponding labels. As a result, a tree structure is built to represent the underlying hierarchically related organization of label concepts, which is shown in \cref{fig:cw_tree}. Two concepts connected in the tree structure means they have parent-child relationship, where the parent is located at the lower hierarchy level. Besides the parent-child relation, we further introduce two notions -- brother and cousin. If two concepts have the same parent, then they are brothers. If the parents of two concepts are brothers, then the two concepts are cousins. According to the laws of inheritance: (1) objects with the parent-child relation should be more similar than those with the uncle-child relation (vertical parent-child relationship); and (2) the traits of brothers should be more similar than cousins (horizontal brother-cousin relationship). An effective model should be able to capture both of the vertical relationship and horizontal relationship, so that the representation of any concept in the latent space should be closer to its parent than uncles, and closer to brothers than cousins. For our HaST-CW model shown in \cref{fig:model_loss}, a new HCW module (\cref{sec:Hierarchical Concept Whitening}) is proposed to preserve the vertical relationship, and a novel SC loss (\cref{sec:Semantic Constraint Loss}) is proposed to preserve the horizontal relationship.

\subsection{Hierarchical Concept Whitening}
\label{sec:Hierarchical Concept Whitening}
The hierarchical concept whitening (HCW) module is one of the key components in the HaST-CW model, which aims to disentangle different label concepts while preserving their underlying hierarchical relationship. Specifically, in this work, the set of label concepts were denoted by $C=\{C_{i}\}_{i=1}^{N_{c}}$, where $C_{i}$ represents the $i^{th}$ concept and $N_{c} = 21$ is the number of concepts. For $C_{i}$, its parent, children, brothers and cousins were denoted as $C_{i.\mathcal{P}}$, $\{C_{i.children}\}$, $\{C_{i.\mathcal{B}}\}$ and $\{C_{i.\mathcal{C}}\}$, respectively. A dataset is denoted as $\mathcal{D} \{ \textbf{x}_i,y_i\} ^n_{i=1}$. We use $\textbf{X}^{C_{i}}=\{\textbf{x}_{j}^{C_{i}}\}_{j=1}^{n_{i}}$ to denote the set of $i^{th}$-class samples labeled by $C_{i}$.

In traditional whitening transformation \cite{chen2020concept}, during the training process, data samples are first fed into the model in mini-batches to obtain the latent representation matrix $\textbf{Z}_{d\times n}$, where $n$ is the mini-batch size and $d$ is the dimension of latent representation. We use ResNet as the model backbone in this work. Then a transformation $\psi$ is applied to decorrelate and standardize $\textbf{Z}_{d\times n}$:
\begin{equation}
     \psi(\textbf{Z})=\textbf{W}(\textbf{Z}-\mu{\textbf{1}_{n\times 1}}^T),
     \label{eq:eq1}
\end{equation}
where $\textbf{W}_{d\times d}$ is the orthogonal whitening matrix, and $\mu=\frac{1}{n}\sum^n_{i=1}\textbf{z}_i$ is the sample mean. A property of representation whitening is that $\textbf{Q}^{\textbf{T}}\textbf{W}$ is still a valid whitening matrix if $\textbf{Q}$ is an orthogonal matrix. We leverage this property for interpretable representation learning. In our model, besides decorrelation and standardization, we expect that the transformed representation of samples from concept $C_i$, namely $\textbf{Q}^{\textbf{T}}\psi(\textbf{Z}^{C_i})$, can align well with the $i^{th}$ axis of latent space. Meanwhile, the underlying hierarchical relationship of concepts should also be preserved in their latent representations. That is, we need to find an orthogonal matrix $\textbf{Q}= [\textbf{q}_{1}, \textbf{q}_{2}, \dots, \textbf{q}_{N_{c}}]$ with two requirements: (1) $\textbf{Z}^{C_i}$ should be most activated by $\textbf{q}_{i}$, i.e., the $i^{th}$ column of $\textbf{Q}$; (2) $\textbf{Z}^{C_i}$ should also be activated by $\{\textbf{q}_c\}$, where $c\in \{C_{i.children}\}$ is the child of concept $C_{i}$. The first constraint makes the representation align together with the corresponding concept dimension, and the second one maintains the vertical parent-child relationship between concepts. To this end, the optimization problem can be formulated as:

\begin{align}
\label{eq:eq2}
\max_{\textbf{q}_1,\dots \textbf{q}_{N_{c}}}&\sum^{N_{c}}_{i=1}[ {\frac{1}{n_i}\textbf{q}^T_i\psi(\textbf{Z}^{C_i})\textbf{1}_{n_i \times1} 
      +  }\nonumber\\&\sum_{c\in \{C_{i.children}\}}{\frac{1}{n_i\times N_{cd}}({\textbf{q}_{c}})^T\psi(\textbf{Z}^{C_i})\textbf{1}_{n_i \times1}}], \\
       & \nonumber s.t. \quad \textbf{Q}^T\textbf{Q}= \textbf{I}_d ,
\end{align}

where $N_{cd} = |\{C_{i.children}\}|$ is the number of child concepts of $C_i$. To solve this optimization problem with the orthogonality constraint, a gradient descent method with the curvilinear search algorithm \cite{wen2013feasible} is adopted. With the whitening matrix $\textbf{W}$ and rotation orthogonality matrix $\textbf{Q}$, HaST-CW can replace any batch normalization layer in deep neural networks. The details of representation whitening for HaST-CW is summarized in Algorithm \ref{alg:forward}.

The overall training pipeline of our HaST-CW model is shown in \cref{alg:HaST-CW}. We adopt an alternative training scheme. In the first stage, the deep neural network is trained with the traditional classification loss. In the second stage, we solve for $\textbf{Q}$ to align representation dimension with semantic concepts. The two stages work alternatively during the training process. The classification  loss of the first stage is defined as: 
\begin{equation}
\min_{\theta,\omega,\textbf{W},\mu,}\frac{1}{m}\sum^m_{i=1}\ell(g(\textbf{Q}^T\psi( \Phi(\textbf{x}_i;\theta);\textbf{W},\mu);\omega);y_i),
\label{eq:eq3}
\end{equation}
where $\Phi(\cdot)$ and $g(\cdot)$ are layers before and after the HaST-CW module parameterized by $\theta$ and $\omega$, respectively. $\psi(\cdot)$ is the whitening transformation parameterized by the sample mean $\mu$ and whitening matrix $\textbf{W}$. The rotation orthogonal matrix $\textbf{Q}$ will be updated according to \cref{eq:eq2} in the second stage. The operation of $\textbf{Q}^T\psi(\cdot)$ forms the HCW module. During the first training stage, $\textbf{Q}$ will be fixed and other parameters ($\theta,\omega,\textbf{W},\mu$) will be optimized according to \cref{eq:eq3} to minimize the classification error. The first stage will take $T_{thre}$ mini batches (we set $T_{thre}=30$ in experiments). After that, $\textbf{Q}$ will be updated by the Cayley transform~\cite{wen2013feasible}: 
\begin{equation}
    \textbf{Q}^\prime = (\textbf{I}+\frac{\eta}{2}\textbf{A})^{-1}(\textbf{I}-\frac{\eta}{2}\textbf{A})\textbf{Q},
    \label{eq:eq4}
\end{equation}
\begin{equation}
    \textbf{A} = \textbf{G}\textbf{Q}^T-\textbf{Q}{\textbf{G}}^T,
    \label{eq:eq5}
\end{equation}
where $\textbf{A}$ is a skew-symmetric matrix. $\textbf{G}$ is the gradient of the concept alignment loss, which is defined in \cref{alg:HaST-CW}. $\eta$ is the learning rate. At the end of the second stage, an updated $\textbf{Q}^\prime$ will participate in the first training stage of the next iteration.

\begin{algorithm}[tb]
   \caption{Forward Pass of HCW Module}
   \label{alg:forward}
\begin{algorithmic}[1]
   \STATE {\bfseries Input:} mini-batch input $\textbf{Z}\in \mathbb{R}^{d\times n}$\
   \STATE {\bfseries Optimization Variables:} orthogonal matrix $\textbf{Q}\in  \mathbb{R}^{d\times d}$\
   \STATE {\bfseries Output:} whitened representations $\hat{\textbf{Z}}\in 	\mathbb{R}^{d\times n}$\
   \STATE The batch mean: $\mu = \frac{1}{n}\textbf{Z}\cdot \textbf{1}$ \ 
   \STATE The centered representations: $\textbf{Z}_\textbf{C} = \textbf{Z}-\mu \cdot \textbf{1}^T$\
   \STATE Calculate ZCA-whitening matrix \textbf{W}\
   \STATE Calculate the whitened representation: $\hat{\textbf{Z}}=\textbf{Q}^T\textbf{W}\textbf{Z}_\textbf{C}$\
\end{algorithmic}
\end{algorithm}

\begin{algorithm*}[tb]
   \caption{The Overall Framework of HaST-CW}
   \label{alg:HaST-CW}
\begin{algorithmic}[1]
    \STATE {\bfseries Input:} Training dataset $\mathcal{D}_T =	\{ \textbf{x}_i,y_i\} ^n_{i=1}$,  Concept datasets $\mathcal{D}_C =\; \{\textbf{X}^{C_1}$, $\textbf{X}^{C_2}$, \dots, $\textbf{X}^{C_{N_{c}}}\}$\
    \STATE {\bfseries Optimization Variables:} $\textbf{W}, \textbf{Q}, \theta, \mu, \omega, \textbf{G} = [\textbf{g}_1, \textbf{g}_2, \dots, \textbf{g}_{N_{c}}]$
    \STATE {\bfseries Hyperparameters:} $\beta, \eta$
   
    \FOR {$t =1\; to\; T$}
        \STATE Randomly sample a mini-batch$ \{ \textbf{x}_i,y_i\} ^n_{i=1} $from $\mathcal{D}_T$
        \STATE Do one step of SGD w.r.t $\theta$ and $\omega$ on the loss $\frac{1}{n}\sum^n_{i=1}\ell(g(\textbf{Q}^T\psi( \Phi(\textbf{x}_i;\theta);\textbf{W},\mu);\omega);y_i)$
        \STATE Update $\textbf{W}$ and $\mu$ by exponential moving average
        \IF{$t$ mod $T_{thre}$ = 0}
            \FOR {$i \in \{1, 2, \dots,N_{c}\}$}
                \STATE Sample mini-batches $\{\textbf{x}_{j}^{C_{i}}, y_{j}\}^{n_{i}}_{j=1}$ from   $\mathcal{D}_C $ 
                \STATE $\textbf{g}_i = -\frac{1}{n_{i}}\sum^{n_{i}}_{j=1} \psi(\Phi(\textbf{x}^{C_i}_j;\theta);\textbf{W},\mu)$ 
                \STATE $C_{child} \in \{C_{i.children}\}$,and $child \in \{1,2,\dots, N_{c}\}$
                \STATE $N_{child} = |\{C_{i.children}\}|$ 
                \FOR {$child$} 
                    \STATE Sample mini-batches $\{\textbf{x}_{j}^{C_{child}}, y_{j}\}^{n_{child}}_{j=1} $
                    \STATE $\textbf{g}_{child} = -\frac{1}{n_{child}\times N_{child}}\sum^{n_{child}}_{j=1} \psi(\Phi(\textbf{x}^{C_{child}}_j;\theta);W,\mu)$
                \ENDFOR
            \ENDFOR
        \ENDIF
    \ENDFOR
\end{algorithmic}
\end{algorithm*}

\subsection{Semantic Constraint Loss}
\label{sec:Semantic Constraint Loss}
Besides preserving the vertical parent-child relationship of concepts, we further model the horizontal relation between concepts that are at the same hierarchy level (i.e., brothers or cousins). Different from the HCW in \cref{eq:eq2} that focuses on concept alignment, here we directly control the distance between representations of different concepts with the horizontal relation~\cite{9chopra2005learning,10schroff2015facenet}. To this end, we propose a Semantic Constraint (SC) loss to model the horizontal brother-cousin relationship as below: 
\begin{align}
&\quad\quad\quad{\mathcal{L}}_{SC} = {\alpha\mathcal{L}}_{\mathcal{B}} + {\beta\mathcal{L}}_{\mathcal{C}}, \label{eq:eq6} \\
{\mathcal{L}}_{\mathcal{B}}=\sum_{j}&\sum_{{\mathcal{B}}_{i}\in{\{C_{i.\mathcal{B}}\}}}\sum_{k} max\{0,m_{\mathcal{B}}-d({\textbf{z}}^{C_{i}}_{j},{\textbf{z}}^{\mathcal{B}_{i}}_{k})\}, \nonumber\\
{\mathcal{L}}_{\mathcal{C}} =\sum_{j}&\sum_{{\mathcal{B}}_{i}\in{\{C_{i.\mathcal{B}}\}}}\sum_{{\mathcal{C}}_{i}\in{\{C_{i.\mathcal{C}}\}}}\sum_{k}\sum_{l} max\{0,d({\textbf{z}}^{C_{i}}_{j},{\textbf{z}}^{\mathcal{B}_{i}}_{k})\nonumber\\-d&({\textbf{z}}^{C_{i}}_{j},{\textbf{z}}^{\mathcal{C}_{i}}_{l})+m_{\mathcal{C}}\}. \nonumber
\end{align}

There are two components in the SC loss and their contributions are controlled by two hyperparameters -- $\alpha$ and $\beta$. The first term ${\mathcal{L}}_{\mathcal{B}}$ is a contrastive loss, which takes a pair of image representations labeled by two brother concepts as input and enlarges the distance between them. It uses a hyperparameter $m_{\mathcal{B}}$ to control the distance. The distance between two concepts increases when $m_\mathcal{B}$ is set larger. ${\mathcal{B}}_{i}\in{\{C_{i.\mathcal{B}}\}}$ denotes one of the brothers of concept $C_{i}$. The second term ${\mathcal{L}}_{\mathcal{C}}$ is a triplet loss.  It takes three inputs: the anchor image representation ${\textbf{z}}^{C_{i}}_{j}$, the image representation ${\textbf{z}}^{\mathcal{B}_{i}}_{k}$ labeled by brother concept of the anchor, and the image representation ${\textbf{z}}^{\mathcal{C}_{i}}_{l}$ labeled by cousin concept of the anchor. ${\mathcal{C}}_{i}\in{\{C_{i.\mathcal{C}}\}}$ denotes the cousins of concept $C_{i}$. The triplet loss encourages the anchor-brother distance to be smaller compared with the anchor-cousin distance in representation space. In this way, the distance of image representations from brother classes tends to be smaller than the distance of image representations from cousin classes. The gap between the two types of distance is controlled by the margin value $m_{\mathcal{C}}$. Consequently, the hierarchical concept whitening module, together with the SC loss, enables the latent representations of concepts with similar semantics to be close with each other in the latent space.

\subsection{Latent Feature Maps Activation}
The proposed HaST-CW model can generate latent representations ($\hat{\textbf{z}}_{i}$) for input images ($\textbf{x}_i$) at each neural network layer by $\hat{\textbf{z}}_{i}=\textbf{Q}^T\psi( \Phi(\textbf{x}_i;\theta);\textbf{W},\mu)$. The latent representation can be used to assess the interpretability of the learning process by measuring the degree of activation of $\hat{\textbf{z}}_{i}$ at different concept dimensions (i.e. $\{\textbf{q}_{i}\}$). In the implementation, $\Phi(\cdot)$ is a CNN based deep network, whose convolution output $\textbf{z}_{i}= \Phi(\textbf{x}_i;\theta)$ is a tensor with the dimension $\textbf{z}_{i} \in R^{d\times h\times w}$. Since $\hat{\textbf{z}}_{i}$ is calculated by $\hat{\textbf{z}}_{i} = \textbf{Q}^T\psi(\textbf{z}_{i})$ where $ \textbf{Q}^T\in R^{d\times d}$, we obtain $\hat{\textbf{z}}_{i} \in R^{d\times h\times w}$, where $d$ is the channel dimension and $h\times w$ is the feature map dimension. The hierarchical concept whitening operation $\textbf{Q}^T\psi(\cdot)$ is conducted upon the $d$ feature maps. Therefore, different feature maps contain the information of whether and where the concept patterns exist in the image. However, as a tensor the feature map cannot directly measure the degree of \textit{concept activation}. To solve this problem and at the same time to reserve both of the high-level and low-level information, we first apply the max pooling on the feature map and then use the mean value of the downsteam feature map to represent the original one. By this way, we reshape the original feature map $\textbf{z}_{i} \in R^{d\times h\times w}$ to $\textbf{z}_{i}^\prime \in R^{d\times 1}$. Finally, $\textbf{z}_{i}^\prime$ is used to measure the activation of image $\textbf{x}_i$ at each concept dimension.


\section{Experiments}

In the experiments section, we first visually demonstrate how our method can effectively learn and hierarchically organize concepts in the latent space (\cref{Visualization-of-Semantic-Map}). We also show that (\cref{Concept-Alignment}), compared to existing concept whitening methods, HaST-CW not only separates concepts, but also can separate groups of semantically related concepts in the latent space. After that, we discuss the advantages offered by our method with quantitative results and intuitive examples (\cref{Interpretable-Image-Classification}) compared with baselines, including the CW module and ablated versions of our method. 

\subsection{Experimental Setting}

\subsubsection{Data Preparation}
In this work, we use a newly collected and curated Agri-ImageNet dataset to evaluate the proposed HaST-CW model. In total, 9173 images from 21 classes are used in our experiments. Each image is labeled with the class at the highest possible hierarchy level. For example, an image of Melrose apple will be labeled as "Melrose" rather than the superclass "Apple". Then we divide images per class into three parts by 60\%/20\%/20\% for a standardized training/validation/test splitting. Because the resolution of the original images can range from 300 to 5000, we adopt the following steps to normalize the image data: 1) we first lock aspect ratio and resize the images to make the short edge to be 256; 2) During each training epoch, the images in the training and validation datasets are randomly cropped into 224×224; 3) During testing process, images in the test dataset are center cropped to be of size 224×224; 4) After cropping, the pixel values of images are normalized to [0,1]. Then, the whole training dataset is divided into two parts ($\mathcal{D}_{T}$ and $\mathcal{D}_{C}$ in \cref{alg:HaST-CW}).  $\mathcal{D}_{C}$ is the concept dataset used to update the matrix $\textbf{Q}$ in the second stage (\cref{eq:eq4}). It is created by randomly selecting 64 images from each class in the training dataset. The remaining images in the training dataset $\mathcal{D}_{T}$ are used in the first stage to train the model parameters (\cref{eq:eq3}). 

\subsubsection{Model Setting}

In this work, we use several ResNet structures \cite{11he2016deep} to extract features from images, including ResNet18 and ResNet50. During the training process, the two-stage training scheme adopts a 30-to-1 ratio to alternatively train the whole framework. In this case, after 30 mini batches of continuous training, the model will pause and the rotation orthogonal matrix $\textbf{Q}$ will be optimized at the next mini batch. Two hyper-parameters $\alpha$ and $\beta$ in the SC loss are set to be 1.0. Adam optimizer is used to train the whole model with a learning rate of 0.1, a batch size of 64, a weight decay of 0.01, and a momentum rate of 0.9.

\subsection{Visualization of Semantic Map} \label{Visualization-of-Semantic-Map}
\begin{figure}[t]
\vskip 0.2in
\begin{center}
\centerline{\includegraphics[width=\columnwidth]{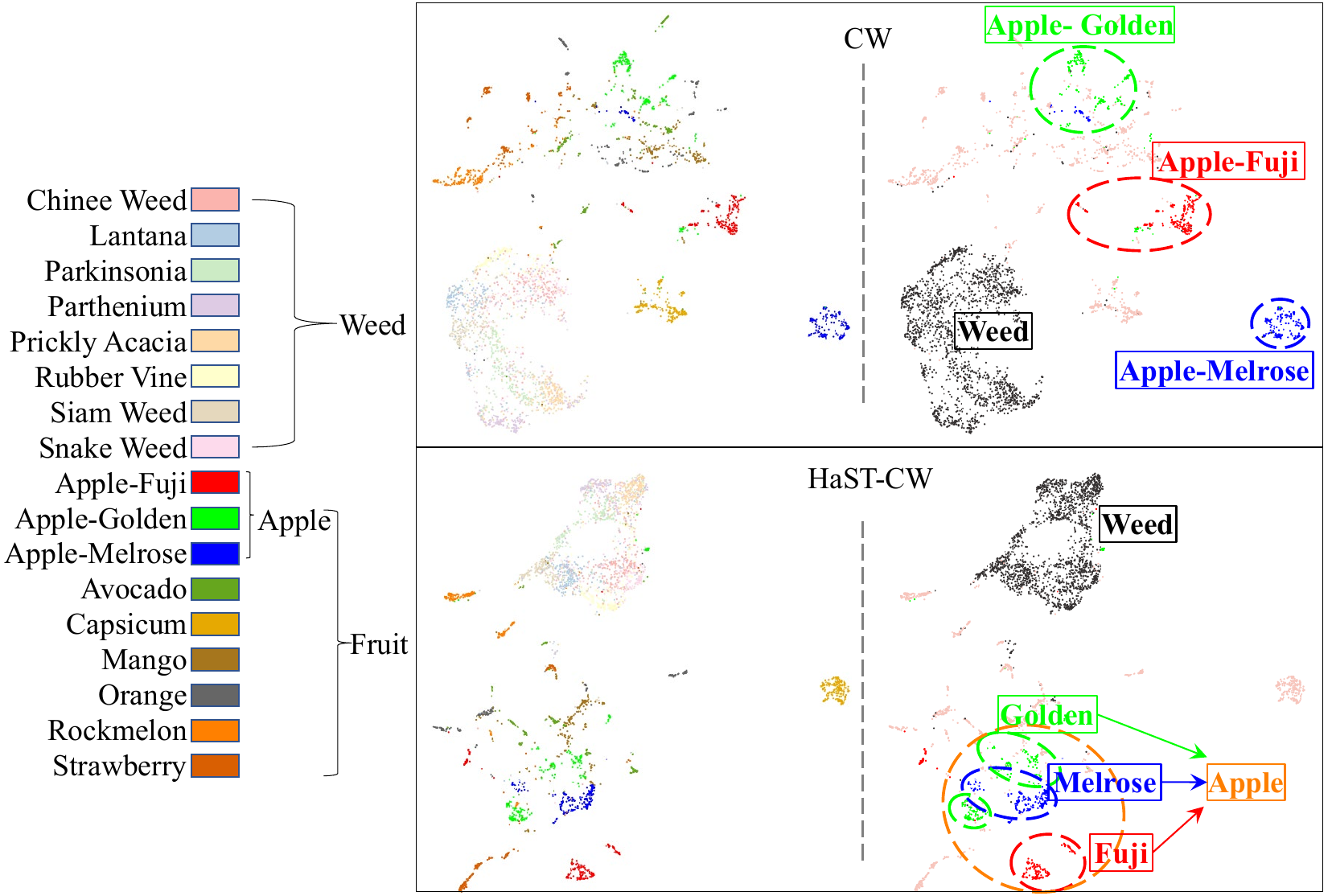}}
\caption{UMAP visualization of the latent space embedding with Agri-ImageNet images, colored according to the legend of image labels on the left.  The top panel shows the results of the CW method and we highlight the super-concept ``Weed" (black) and three sub-concepts.  As shown in the bottom panel, we apply the same rules to the output of HaST-CW and visualize the results.   In addition, we draw an orange circle that encapsulates three types of apples to represent the super-concept ``Apple".   }
\label{main}
\end{center}
\vskip -0.2in
\end{figure}

To illustrate the learned semantic hierarchical structure, we show the representations extracted from the latent hidden layer of all the samples in \autoref{main}. For better visualization, we use Uniform Manifold Approximation and Projection (UMAP) ~\cite{mcinnes2018umap} to project the representations to a two-dimensional space. All the images are color coded using the 17 sub-concepts which are defined on the left of \autoref{main}. The top panel shows the result using CW method. In general, all the concepts are assembled as small groups, but neither semantic relations nor hierarchical structures have been learned. We highlight the super-concept of ``Weed" (black) and three sub-concepts ( ``Apple Golden" - green, ``Apple Fuji" - red and ``Apple Melrose" - blue) in the right column. We can see that the three types of apple (sub-concepts) are evenly distributed along with other fruits samples. The bottom panel shows our HaST-CW results. All the different concepts successfully keep their distinct cluster patterns as CW result. After our two-stage training process to instill the semantic and hierarchical knowledge, the three types of apple images have been pulled together and form a new concept (``Apple" with orange circle) at a higher level. Moreover, the newly learned concept of ``Apple" simultaneously possesses sufficient distance to ``Weed" (different super-concept) and maintains relatively close relations to ``Strawberry", ``Orange", ``Mango" as well as other types of ``Fruit". This result demonstrates the effectiveness of our hierarchical semantic concept learning framework, without negatively affecting the overall classification performance. 

\subsection{Efficiency and Accuracy of Concept Alignment} \label{Concept-Alignment}
\begin{figure}[t]
\vskip 0.2 in
\begin{center}
\centerline{\includegraphics[width=\columnwidth]{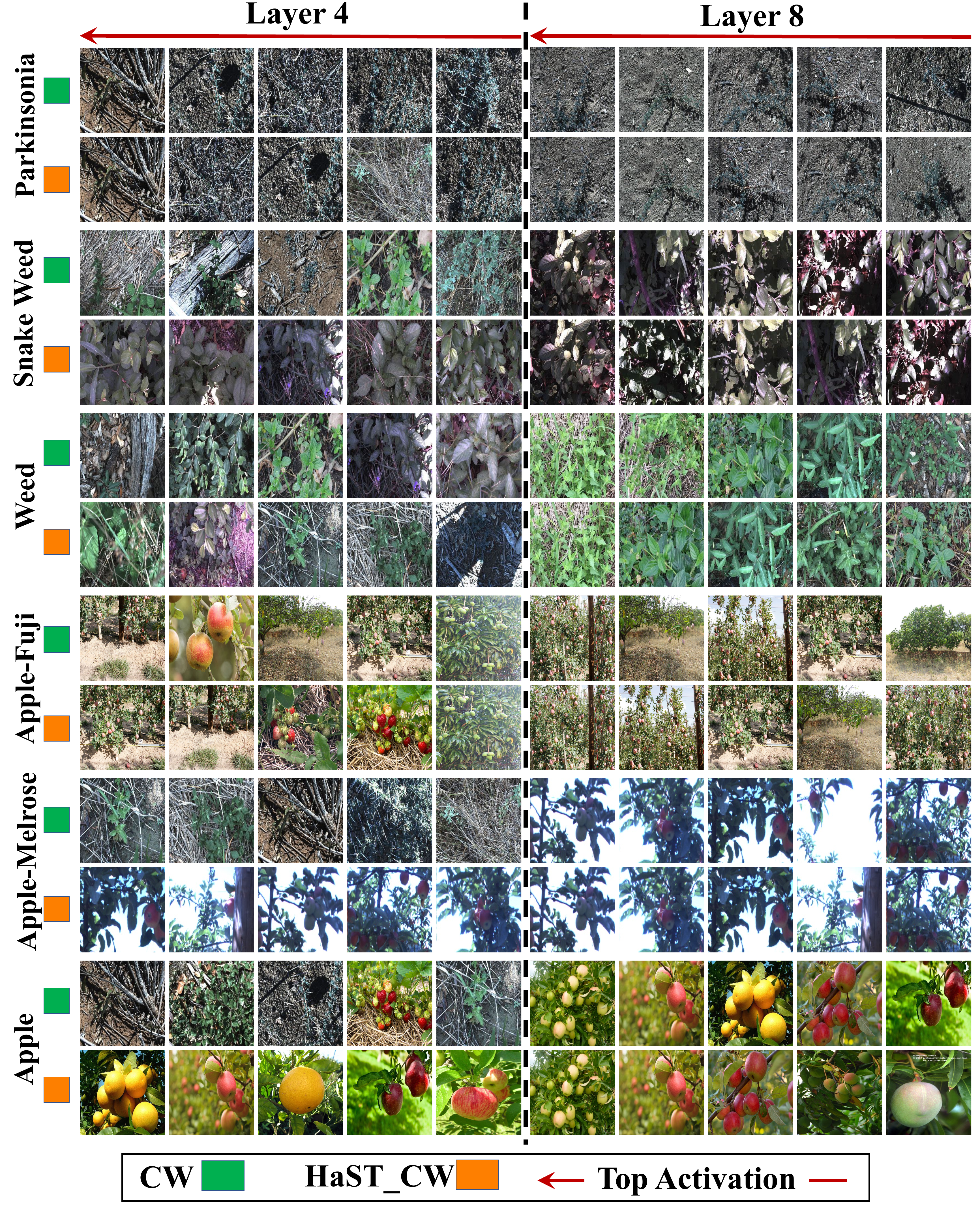}}
\caption{Top 5 activation images of each concept. The image panel is divided into two sets of columns: the left set of columns contains the results of layer 4  (a shallow layer), whereas the right set of columns holds the results of layer 8 (a deeper layer). Each concept covers two rows that correspond to the results of the conventional CW (marked by green boxes) and the proposed HaST-CW (marked by orange boxes), respectively. }
\label{ht}
\end{center}
\vskip -0.2in
\end{figure}

In this section, we compare the learning efficiency and accuracy of the proposed HaST-CW with that of the conventional CW method. We track the alignment between image representations and their corresponding concepts at each layer. Specifically, we randomly select six concepts, and for each concept we sort and select the top five images whose representations show the strongest activation at the corresponding concept axis. We show the results at both shallow and deep layers (layer 4 vs. layer 8) in \autoref{ht}. From the results of layer 4 (the left column) we can see that most of the top five images obtained by conventional CW (the rows marked by green box) are mismatched with the corresponding concepts. For example, the five images under the concept of ``Apple-Melrose" obtained by CW are from the ``Weed" class. The five images under the concept of ``Snake Weed" are actually from other subclass of ``Weed". Moreover, this situation continues in the following layers and has not been changed until layer 8. On the contrary, with the help of our designed semantic constraint loss, our HaST-CW (the rows marked by orange boxes) can learn the intrinsic concept faster and achieves the best performance at an earlier training stage (e.g., at a shallow layer). This result demonstrates that by paralleling multiple HCW layers the proposed HaST-CW model can capture the high-level features more efficiently.

\begin{figure}[t]
\vskip 0.2 in
\begin{center}
\centerline{\includegraphics[width=1.0\columnwidth]{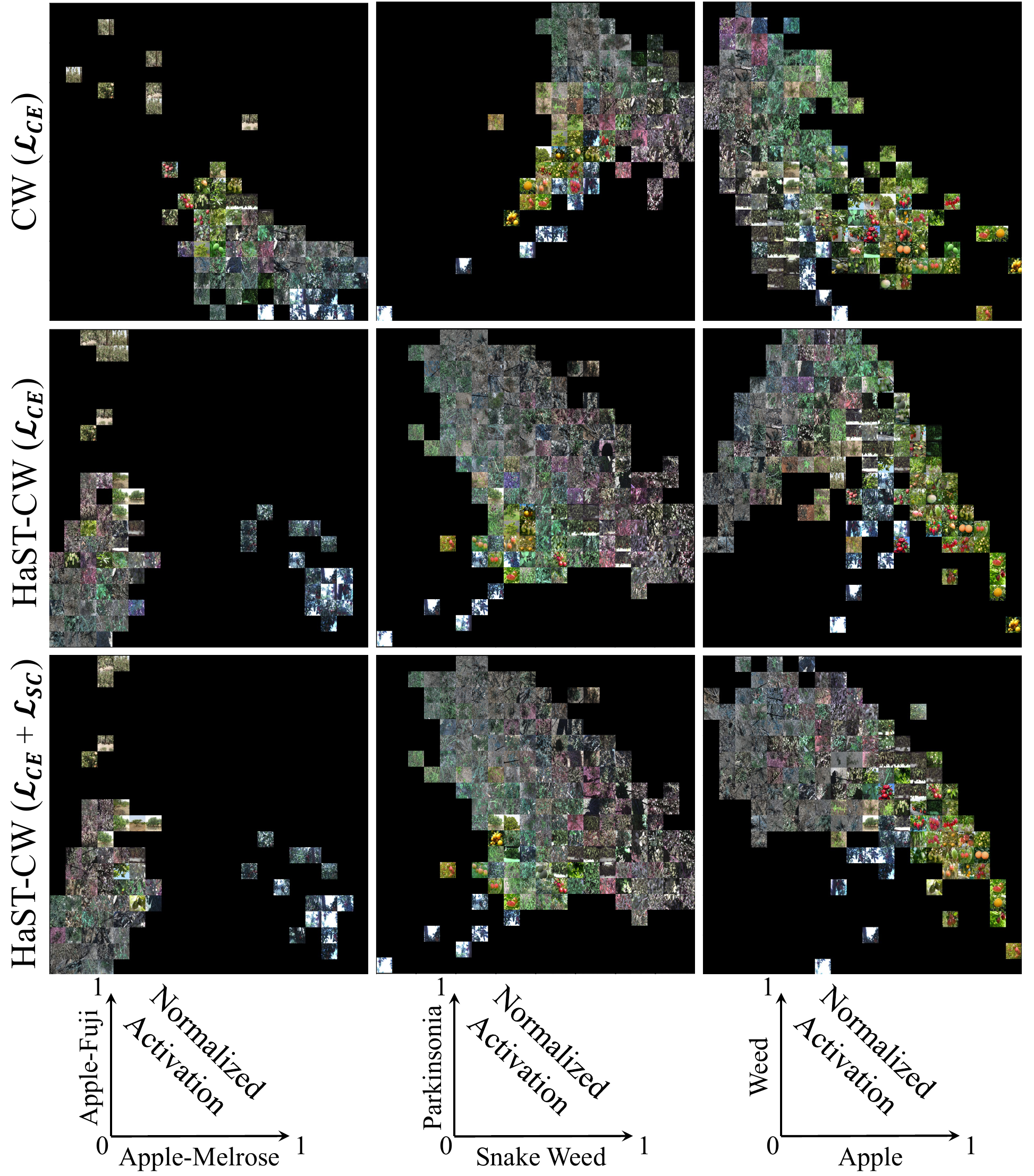}}
\caption{Data distribution in the concept latent space. Three pairs of concepts corresponding to different semantic hierarchy levels are selected. For each concept pair, a two-dimensional space is built by taking the concepts as axes. To visualize the alignments between images and the concepts, the images are projected into the two-dimensional space by similarity values between image representations and the two concept representations. Different rows in the figure panel correspond to different methods and the concept axes (space) are defined at the bottom.}
\label{representation}
\end{center}
\vskip -0.2in
\end{figure}

To further demonstrate the alignment between images and the corresponding concepts, we project each image in the test dataset into a latent space where each concept can be represented by an axis. To visualize the alignments at different concept hierarchies (\autoref{fig:cw_tree}), we show three pairs of concepts which belong to different hierarchical levels as examples: ``Apple-Melrose"-``Apple-Fuji" is from hierarchy 3 (H-3), ``Snake Weed"-``Parkinsonia" is from hierarchy 2 (H-2), and ``Weed"-``Apple" crosses hierarchies 1 and 2 (``Weed": H-1, ``Apple": H-2). Within each concept pair, a two-dimensional space has been built by taking the two concepts as axes. Thus, each image can be mapped into the space by calculating the similarity between image representation and the two concept representations. The results are shown in \autoref{representation}. Different rows correspond to different methods and the concept axes (space) are defined at the bottom. 

The first column of \autoref{representation} shows the data distribution in the two-dimensional space of ``Apple-Melrose"-``Apple-Fuji" concept pair. The images belonging to Apple-Melrose class should have the highest similarity with the concept of ``Apple-Melrose", and thereby they should be located at the right-bottom corner. Similarly, the images of Apple-Fuji class should be located at the left-top corner. The other images should distribute in the space according to the similarity with the two concepts. For example, compared to images of fruit-related classes, images of weed-related classes will have lower semantic similarity with the two concepts, so they should locate near the origin point (left-bottom corner). As shown in the first column, the two models which adopt the HaST-CW method (the second and third rows) can better follow the above-mentioned patterns. While in the CW model (the first row), nearly all the images are gathered at the right-bottom corner. This may be due to the high similarity between the two concepts considered, since they share the same super-class of ``Apple". As a result, CW model may be limited in distinguishing different classes with high semantic similarity. A similar situation happens in the second column with the concept pair of ``Snake Weed"-``Parkinsonia". These results suggest that compared to CW method, HaST-CW can better capture the subtle differences of semantic-related classes. 

The third column shows the results of the concept pair of two super-classes: ``Weed" and ``Apple". As each of the super-class concept contains multiple sub-classes, the intra-class variability is greater. Our proposed HaST-CW, together with the SC loss (the third row), can effectively capture the common visual features and project the ``Weed" and ``Apple" images to the left-top and right-bottom, respectively. At the same time, the images belonging to different sub-classes under ``Weed" and ``Apple" are assembled as blocks instead of scattered along the diagonal line. In the other two methods, especially in the CW method (the first row), the images of ``Weed" class spread out over a wide range along the vertical axis. This result suggests that the proposed HaST-CW with SC loss can effectively model both the inter- and intra- class similarity.

\subsection{Interpretable Image Classification} 
\label{Interpretable-Image-Classification}
In this section, we compare the classification performance of the proposed HaST-CW method and the SC loss function with the conventional CW method using different backbones: ResNet18 and ResNet50. The results are summarized in \Cref{table:acc-table}. Different rows correspond to different model settings. Within each model setting, we repeat the experiments for five times to reduce the effect of random noise. The mean and variance of accuracy (ACC.) are reported in the fourth column. From the results, we can see that the classification performance is slightly better than the other three model settings. This result indicates that the proposed HaST-CW model can improve the interpretability without hurting predictive performance.   

\begin{table}[th]
\caption{ Comparison of Classification Performance.}
\label{table:acc-table}
\vskip 0.15in
\begin{center}
\begin{small}
\begin{sc}
\begin{tabular}{cccc}
\toprule
Module  & Backbone & Loss    & Acc.       \\ \midrule
CW      & ResNet18 & ${\mathcal{L}}_{CE}$     & 63.48 $\pm$ 0.68 \\
CW      & ResNet50 & ${\mathcal{L}}_{CE}$     & 69.25 $\pm$ 3.93 \\
HaST-CW & ResNet50 & ${\mathcal{L}}_{CE}$     & 69.30 $\pm$ 3.75 \\
HaST-CW & ResNet50 & ${\mathcal{L}}_{CE}+{\mathcal{L}}_{SC}$ & \textbf{69.49} $\pm$ \textbf{3.20} \\ \bottomrule
\end{tabular}
\end{sc}
\end{small}
\end{center}
\vskip -0.1in
\end{table}

\begin{figure}[t]
\vskip 0.2in
\begin{center}
\centerline{\includegraphics[width=0.85\columnwidth]{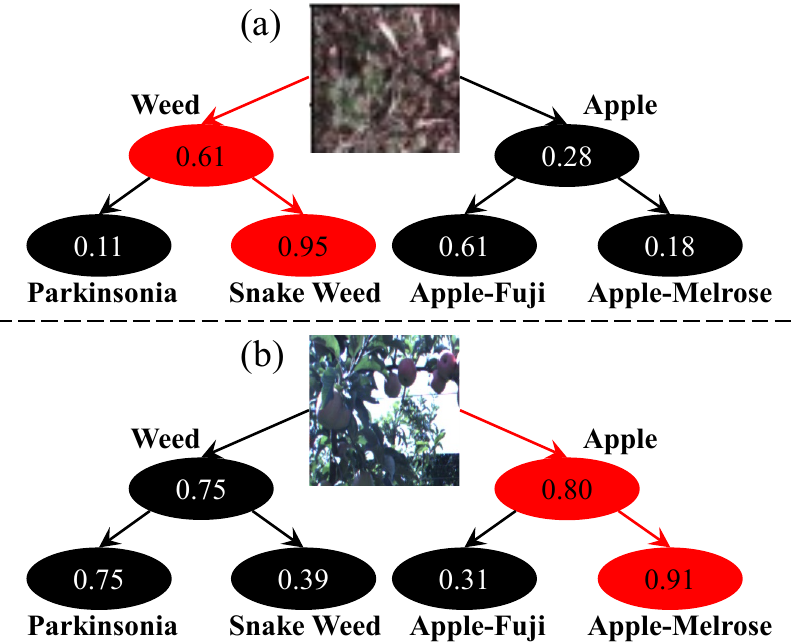}}
\caption{Hierarchical activation tree. We randomly select two images from the Apple-Melrose class and the Snake Weed class. For each image, activation values corresponding to the 6 concepts are calculated and normalized to [0, 1]. The highest activation values are highlighted with red along the hierarchical path. }
\label{tree-represent}
\end{center}
\vskip -0.2in
\end{figure}

To track and visualize the classification process, we randomly select two images from Apple-Melrose class and Snake Weed class. The activation values between each image with the six relevant concepts are calculated and normalized to [0, 1]. The images, concepts and activation values are organized into a hierarchical activation tree. The results are shown in \autoref{tree-represent}. We could observe that the activation values of each image correctly represent the semantic relationship between the images and the concepts. For example, in \autoref{tree-represent} (a), the image located at the root is from Snake Weed class which is a subclass of Weed. The activation values of the image are consistent with this relationship and possess the highest activation values on the two concepts -- ``Weed" and ``Snake Weed".

\section{Conclusion and Future Work}
In this study, we propose a new HaST-CW and demonstrate its superiority over Concept Whitening ~\cite{chen2020concept}. HaST-CW decorrelates representations in the latent space and aligns concepts with corresponding dimensions. In addition, it correctly groups concepts at different granularity levels in the latent space and preserves hierarchical structures of concepts of interest. By doing so, we can interpret concepts better and observe the semantic relationships among concepts. 

We believe there are many possibilities for future work. One promising direction is automatically learning concepts from data. In this scenario, we can jointly learn possible concepts from common abstract features among images and how to represent these learned concepts in the latent space. For example, it might be possible to develop unsupervised or weakly-supervised methods to automatically learn the concept tree from data. By jointly learning concepts, their representations, and relations, the model may discover more data-driven semantic structures.

HaST-CW can also be extended with post-hoc interpretability strategies (such as saliency-based methods that highlight focused areas used for classification). Such explanations at the concept level can provide a more global view of model behaviors.

In addition, while this work focuses on the the natural image domain, the idea of leveraging hierarchical knowledge to guide representation learning is generalizable to other domains such as natural language processing~\cite{rezayi2023exploring,liu2023summary,cai2022coarse} and medical image analysis~\cite{li2023artificial,bi2023community,liu2022survey,wang2023review,zhang2023segment,zhang2023differentiating,qiang4309357deep,dai2022graph,liu2022discovering}. Exploring knowledge-infused learning in different domains~\cite{zhao2023brain,holmes2023evaluating,liu2023context,ma2023impressiongpt,liu2023radiology,ding2023deep,ding2022accurate} and tasks~\cite{liao2023differentiate,liu2023deid,liao2023mask,wu2023exploring,zhong2023chatabl}, including innovative applications~\cite{dai2023samaug,zhang2023beam,dai2023chataug}, is an interesting future direction.

In conclusion, as deep learning models become increasingly complex, model interpretability is crucial for understanding behaviors, gaining trust, and enabling human-AI collaboration. Our work complements previous work and lays a solid foundation for further exploration.



\bibliography{ref_nh}
\bibliographystyle{IEEEtran}

\newpage

\end{document}